\documentclass[sigconf]{acmart}
\AtBeginDocument{%
  }

\copyrightyear{2025} 
\acmYear{2025} 
\setcopyright{acmlicensed}\acmConference[SIGIR '25]{Proceedings of the 48th
International ACM SIGIR Conference on Research and Development in
Information Retrieval}{July 13--18, 2025}{Padua, Italy}
\acmBooktitle{Proceedings of the 48th International ACM SIGIR Conference on
Research and Development in Information Retrieval (SIGIR '25), July 13--18,
2025, Padua, Italy}
\acmDOI{10.1145/3726302.3729987}
\acmISBN{979-8-4007-1592-1/2025/07}

\usepackage{bbm}
\usepackage{subfigure}
\usepackage{multirow}
\usepackage{algorithm,algorithmic}
\usepackage{enumitem}
\usepackage{color,xspace}
\usepackage{xcolor}
\usepackage{soul,ulem}
\usepackage{amsmath}
\usepackage{bbm}
\usepackage{colortbl}

\usepackage{tcolorbox}  % 文本加框
\tcbuselibrary{breakable}   % 配合文本加框，使得框可以跨页

\newcommand{\name}{GAVE\xspace}

\begin{document}

\title{Generative Auto-Bidding with Value-Guided Explorations}

\author{Jingtong Gao}
\affiliation{%
  \institution{City University of Hong Kong}
  \city{Hong Kong}
  \country{China}}
\email{jt.g@my.cityu.edu.hk}

\author{Yewen Li}
\affiliation{%
  \institution{Nanyang Technological University}
  \city{Singapore}
  % \state{Beijing Shi}
  \country{Singapore}
  }
\email{yewen001@e.ntu.edu.sg}

\author{Shuai Mao}
\affiliation{%
  \institution{The Chinese University of Hong Kong}
  \city{Hong Kong}
  \country{China}}
\email{smao@mae.cuhk.edu.hk}

\author{Peng Jiang}
\affiliation{%
  \institution{Kuaishou Technology}
  \city{Beijing}
  \country{China}
}
\email{jiangpeng07@kuaishou.com}

\author{Nan Jiang}
% \authornote{Corresponding author.}
\affiliation{%
  \institution{Kuaishou Technology}
  \city{Beijing}
  \country{China}
}
\email{jiangnan07@kuaishou.com}

\author{Yejing Wang}
\affiliation{%
  \institution{City University of Hong Kong}
  \city{Hong Kong}
  \country{China}}
\email{yejing.wang@my.cityu.edu.hk}

\author{Qingpeng Cai}
\authornote{Corresponding authors.}
\affiliation{%
  \institution{Kuaishou Technology}
  \city{Beijing}
  \country{China}
}
\email{caiqingpeng@kuaishou.com}

\author{Fei Pan}
% \authornote{Corresponding author.}
\affiliation{%
  \institution{Kuaishou Technology}
  \city{Beijing}
  \country{China}
}
\email{panfei05@kuaishou.com}

\author{Peng Jiang}
% \authornote{Corresponding author.}
\affiliation{%
  \institution{Kuaishou Technology}
  \city{Beijing}
  \country{China}
}
\email{jiangpeng@kuaishou.com}

\author{Kun Gai}
\affiliation{%
  \institution{Unaffiliated}
  \city{Beijing}
  \country{China}
}
\email{gai.kun@qq.com}

\author{Bo An}
\affiliation{%
  \institution{Nanyang Technological University}
  \city{Singapore}
  \country{Singapore}
  }
\email{boan@ntu.edu.sg}

\author{Xiangyu Zhao}
\authornotemark[1]
\affiliation{%
  \institution{City University of Hong Kong}
  \city{Hong Kong}
  \country{China}}
\email{xianzhao@cityu.edu.hk}

\renewcommand{\shortauthors}{Gao et al.}

\begin{abstract}

Auto-bidding, with its strong capability to optimize bidding decisions within dynamic and competitive online environments, has become a pivotal strategy for advertising platforms. Existing approaches typically employ rule-based strategies or Reinforcement Learning (RL) techniques. However, rule-based strategies lack the flexibility to adapt to time-varying market conditions, and RL-based methods struggle to capture essential historical dependencies and observations within Markov Decision Process (MDP) frameworks. Furthermore, these approaches often face challenges in ensuring strategy adaptability across diverse advertising objectives. 
Additionally, as offline training methods are increasingly adopted to facilitate the deployment and maintenance of stable online strategies, the issues of documented behavioral patterns and behavioral collapse resulting from training on fixed offline datasets become increasingly significant.
To address these limitations, this paper introduces a novel offline \textbf{G}enerative \textbf{A}uto-bidding framework with \textbf{V}alue-Guided \textbf{E}xplorations (\textbf{\name}). \name accommodates various advertising objectives through a score-based Return-To-Go (RTG) module. Moreover, \name integrates an action exploration mechanism with an RTG-based evaluation method to explore novel actions while ensuring stability-preserving updates. A learnable value function is also designed to guide the direction of action exploration and mitigate Out-of-Distribution (OOD) problems.
Experimental results on two offline datasets and real-world deployments demonstrate that \name outperforms state-of-the-art baselines in both offline evaluations and online A/B tests. By applying the core methods of this framework, we proudly secured first place in the NeurIPS 2024 competition, `AIGB Track: Learning Auto-Bidding Agents with Generative Models'~\footnote{https://tianchi.aliyun.com/competition/entrance/532236/rankingList}. The implementation code is publicly available to facilitate reproducibility and further research~\footnote{https://github.com/Applied-Machine-Learning-Lab/GAVE}.

\end{abstract}





\maketitle

\section{Introduction}
Bidding remains a fundamental component of modern online advertising platforms~\cite{zhao2018deep, zhao2020jointly,gao2024smlp4rec,gao2024hierrec}, enabling businesses to engage target audiences and increase sales. As digital advertising evolves, traditional manual bid management methods have become increasingly inadequate and cost-ineffective, failing to address the demands of today's dynamic market~\cite{jha2024optimizing,liu2024sequential}. The complexity of modern advertising systems, characterized by fluctuating market conditions and diverse user behaviors~\cite{aggarwal2024auto}, requires automated solutions that can adapt to these variations while aligning with advertisers' diverse objectives~\cite{borissov2010automated,wang2023multi}. This need is further intensified by the massive volume of ad impressions requiring real-time processing, where human intervention becomes impractical and often suboptimal for achieving advertising goals~\cite{borissov2010automated, guo2024generative}.   

To meet these requirements, existing solutions have evolved into two primary categories: predefined rule-based strategies~\cite{chen2011real,yu2017online} and Reinforcement Learning (RL)-based methods~\cite{ye2019deep, cai2017real,gao2023autotransfer,zhao2021dear}. While rule-based strategies are computationally lightweight and straightforward to deploy, their static nature makes them ill-suited for dynamic markets and unable to accommodate advertisers' diverse demands~\cite{guo2021reinforcement, cai2017real,liu2022rating,li2023strec}. RL-based approaches, though employing Markov Decision Processes (MDPs)~\cite{ye2019deep,zhao2019deep,liu2023multi,zhao2018recommendations} to adapt to environmental changes and obtain better performance, face a critical structural constraint: the MDP framework’s state-independence assumption inherently disregards temporal dependencies and observations within bidding sequences~\cite{chen2021decision,li2024gas,li2022gromov}. This limitation obstructs the identification of evolving behavioral patterns and market fluctuations, substantially undermining RL’s real-world applicability in highly volatile real-time bidding environments. 

% To address these limitations, we propose the use of Decision Transformer (DT)~\cite{chen2021decision}, a generative approach for offline auto-bidding modeling. By adopting an offline training paradigm, DT circumvents the risks and implementation challenges of online training, ensuring broader applicability across diverse advertising scenarios~\cite{fujimoto2019off}. The generative modeling foundation of DT further enables explicit capture of temporal dependencies and historical bidding context, allowing adaptive decision-making that aligns with the dynamic nature of real-world advertising environments.
Recently, Decision Transformer (DT)~\cite{chen2021decision,gao2025samplellm,gao2025llm4rerank} has emerged as a powerful framework, effectively capturing temporal dependencies and historical context. Therefore, applying DT to offline bidding modeling offers a promising direction for improving strategies. Specifically, by adopting an offline training paradigm, DT circumvents the risks and implementation challenges of online training, ensuring broader applicability across diverse scenarios~\cite{fujimoto2019off,liu2020automated,wang2023plate,li2023hamur}. The generative modeling foundation of DT further enables explicit capture of temporal dependencies and historical bidding context, allowing adaptive decision-making that aligns with the dynamic nature of real-world advertising environments.

However, several critical challenges emerge when implementing a DT-based auto-bidding approach in real-world scenarios. First, practical deployment requires accommodating complex advertising objectives where evaluation metrics extend beyond elementary indicators like total clicks or conversions. These objectives typically involve sophisticated functions with different preferences on interdependent parameters---such as Cost Per Action (CPA) thresholds and Cost Per Click (CPC) ceilings~\cite{guo2024generative,li2024gas}---requiring DT modeling with adaptive optimization objectives to align with diverse operational criteria. Second, directly training DT models in offline environments may restrict to documented behavioral patterns~\cite{kostrikovoffline,lin2023autodenoise} and suffer from behavioral collapse~\cite{fujimoto2019off}, which requires amplified action explorations with stable updates.

To tackle these challenges, we propose a unified framework that enhances DT for offline \textbf{G}enerative \textbf{A}uto-bidding through \textbf{V}alue-guided \textbf{E}xplorations (GAVE). 
First, to accommodate complex advertising objectives~\cite{he2021unified,guo2024generative}, we design \textbf{a score-based Return-To-Go (RTG)} module with customizable score functions, enabling adaptive modeling of various objective requirements like CPA constraints through differentiable programming. 
Second, \textbf{an action exploration mechanism} is proposed alongside an RTG-based evaluation method to explore and evaluate actions outside the fixed dataset while ensuring stability-preserving updates between explored and original actions.
However, it is rather challenging to learn a beneficial strategy through random explorations and avoid Out-of-Distribution (OOD) risks in such a sensitive bidding environment with large action space~\cite{xuoffline,kostrikov2021offline,liu2023exploration}.
Thus, we introduce \textbf{a learnable value function}~\cite{kostrikovoffline,szepesvari1999unified} to guide the action exploration process, directing exploration toward potentially optimal actions. This mechanism anchors explorations within plausible regions while enabling controlled extrapolation, thereby facilitating strategy improvement and further mitigating OOD issues.

Our contributions are summarized as follows:  

\begin{itemize}[leftmargin=*]  
    \item We introduce an innovative framework \name that leverages DT to optimize auto-bidding strategies, which is designed for seamless adaptability to various real-world scenarios.
    \item
    This paper presents three technical innovations: (1) A score-based RTG module with customizable functions for various advertising objectives through differentiable programming; (2) An action exploration mechanism with RTG-based evaluation to ensure stability-preserving updates; (3) A learnable value function to anchor exploration to plausible regions, thus mitigating OOD risks and enabling controlled extrapolation for strategy improvement.

    \item Experiments on two public datasets, along with results from online deployments, demonstrate the effectiveness of \name compared to various state-of-the-art offline bidding baselines. Additionally, by applying the core methods of this framework, we proudly secured first place in the NeurIPS 2024 competition, `AIGB Track: Learning Auto-Bidding Agents with Generative Models' hosted by Alimama~\footnotemark[1].
\end{itemize}  

%By integrating these solutions, our framework not only improves the performance, adaptability, and robustness of auto-bidding strategies but also ensures their practical applicability across diverse and dynamic real-world advertising environments.  

\section{Preliminary}

In this section, we first illustrate the auto-bidding problem, and then introduce the DT-based decision-making process for modeling.

\subsection{Auto-Bidding Problem}~\label{sec:problem}
Consider a sequence of $I$ impression opportunities arriving over a discrete time period $i = 1,...,I$. Advertisers engage in real-time competition by submitting bids $\{b_i\}_{i=1}^I$ for these impressions. 

The auction mechanism operates under the following rules: An advertiser wins impression $i$ if his bid $b_i$ exceeds $b_i^-$, the highest competing bid from other participants. The winning advertiser then incurs a cost $c_i$, which is determined by the auction mechanism. Following standard industry practice~\cite{edelman2007internet,aggarwal2009general}, we adopt the generalized second-price auction mechanism where the winning cost equals the second-highest bid. The advertiser's objective is to maximize the total acquired value through won impressions during the period. This optimization problem can be formally expressed as:
\begin{equation}
\max \sum_{i=1}^{I} x_i v_i
\end{equation}
where $v_i \in \mathbb{R}^+$ represents the advertiser's private valuation for impression $i$ such as conversion or click-through rate, and $x_i \in \{0,1\}$ denotes the binary decision variable indicating auction outcome:
\begin{equation}
x_i = \begin{cases}
1 & \text{if } b_i > b_i^- \\
0 & \text{otherwise}
\end{cases}
\end{equation}

Simultaneously, advertisers must satisfy multiple constraints to ensure efficient campaign management. The fundamental constraint is the total budget limitation:
\begin{equation}\label{equ:budget}
\sum_{i=1}^{I} x_i c_i \leq B
\end{equation}
where $B \in \mathbb{R}^+$ represents the advertiser's total budget.
% For other constraints, among various Key Performance Indicators (KPI), \w{this paper concentrates on Cost Per Action (CPA)} requirements due to their operational significance in performance marketing\w{cite}. The CPA constraint can be formulated as follows:
% \begin{equation}
% \frac{\sum_{i=1}^I x_i c_i}{\sum_{i=1}^I x_i v_i} \leq C
% \end{equation}
% where $C \in \mathbb{R}^+$ denotes the maximum allowable CPA. This ratio quantifies the efficiency of advertising expenditure relative to value creation. Since most of the other KPI constraints can be modeled through a similar expression. \w{In this paper, for the simplicity of expressions, we only consider CPA constraint for modeling.}
Other Key Performance Indicator (KPI) constraints, exemplified by Cost Per Acquisition (CPA), can be formulated as follows:
\begin{equation}
\frac{\sum_{i=1}^I x_i c_i}{\sum_{i=1}^I x_i v_i} \leq C
\end{equation}
where $C \in \mathbb{R}^+$ denotes the maximum allowable CPA. This ratio quantifies the efficiency of advertising expenditure relative to value creation. Since most other KPI constraints can be modeled similarly, we consider only the CPA constraint for simplicity in this paper.
However, unlike budget constraints, which are directly managed by the auction platform, these KPI constraints are generally not strict in practical scenarios. This is because calculating these constraints requires the advertiser's $v_i$ for all bidding impressions, making it possible to determine the true CPA only after the entire bidding process concludes. Nevertheless, we still hope to use them as soft constraints in modeling.

Therefore, the whole bidding process could be expressed as:
\begin{equation}\label{equ:problem1}
\begin{aligned}
\max _{b_1, \cdots, b_I} & \sum_i x_i   v_i \\
\text { s.t. } & \sum_i x_i   c_i \leq B \\
& \frac{\sum_i x_i   c_i}{\sum_i x_i   v_i} \leq C 
\end{aligned}
\end{equation}

Solving this optimization problem presents inherent challenges stemming from both the high cardinality of impressions and the fundamental uncertainty about future auction performance. Previous research~\cite{he2021unified} reformulates this problem as a Linear Programming problem to yield a simplified optimal bidding strategy:

\begin{equation}\label{equ:bequ}
b_i^*=\lambda_0^* v_i-\Sigma_j \lambda_j^*\left(\mathbbm{q}_{i j}\left(1-\mathbbm{1}_{C R_j}\right)-\mathbbm{k}_j \mathbbm{p}_{i j}\right)
\end{equation}
where $b_i^*$ denotes the theoretically optimal bid for impression $i$, $\mathbbm{q}_{i j}$ can be any performance indicator or constant and $\mathbbm{1}_{C R_j}$ is the indicator function of whether constraint $j$ is cost-related. $\mathbbm{p}_{i j}$ and $\mathbbm{k}_j$ can be treated as expanded expressions of $v_i$ and $C$ in Equation~\eqref{equ:problem1} under multiple KPI consditions. This reformulation transforms the auto-bidding problem into the identification of the optimal $\lambda_0^*$ and $\lambda_j^*$ that satisfy all constraints. By substituting Equation~\eqref{equ:bequ} into Equation~\eqref{equ:problem1} {with $j=1$, $\mathbbm{1}_{C R_j}=1$, $\mathbbm{p}_{i j}=v_i$, $\mathbbm{k}_j=C$ and $\mathbbm{q}_{i j}$ being any performance indicator or constant, we can obtain:

\begin{equation}\label{equ:bi}
b_i^* = (\lambda_0^*+\lambda_1^*C) v_i = \lambda^* v_i
\end{equation}
where $\lambda^*=\lambda_0^*+\lambda_1^*C$ serves as the unified bidding parameter. Therefore, many recent studies have sought to address the bidding problem by iteratively identifying the optimal $\lambda^*$ within the bidding process~\cite{he2021unified,li2024gas,su2024a}.
Additionally, it is worth noting that when solving a bidding problem according to Equation~\eqref{equ:bi}, the first condition, i.e., $\text { s.t. }  \sum_i x_i   c_i \leq B$, is always satisfied. This is because the bidding platform will automatically control $x_i$ when the advertiser's budget is insufficient to ensure that the advertiser does not owe money. 
However, the second condition is not always satisfied since there's a gap between our predicted $\lambda$ and the optimal $\lambda^*$. A simple solution to this problem is to add a penalty term about the CPA condition to the objective function in Equation~\eqref{equ:bi} in the evaluation stage for model selection~\cite{su2024a}, which will be discussed later in Section~\ref{sec:opt}.

\subsection{DT-based Auto-bidding}\label{sec:decision}
% \subsection{DT-based Decision Making Process}\label{sec:decision}

% \w{The temporal arrival of impressions prevents the direct computation of $\lambda^*$ as per Equation~\ref{equ:bi}. 
To solve the auto-bidding problem, existing approaches employ either rule-based policies~\cite{balakrishnan2014real,sayedi2018real} or RL methods~\cite{wu2018budget,yuan2022actor} for optimization. However, rule-based policies often fail to adapt to the highly dynamic nature of real-world bidding environments~\cite{he2021unified}, and RL approaches~\cite{wang2022deep} rely on state transitions defined by $s_{t+1} = f(s_t, a_t)$, which complicates the modeling of essential temporal dependencies and historical observations inherent in auction ecosystems~\cite{yuan2013real}.

Recent advancements in transformer architectures~\cite{lin2022survey,kumar2024large} have led to the emergence of DT~\cite{chen2021decision,zheng2022online,wu2024elastic}, positioning them as state-of-the-art for sequential decision-making. DTs excel in capturing long-range dependencies, making them ideal for bidding environments where auction outcomes display significant temporal correlations. Building on this framework, we approach auto-bidding as a sequence modeling task~\cite{su2024a,li2024gas} under DT settings. The bidding period is divided into discrete time steps, with each step configured under specific environmental settings:

\begin{itemize}[leftmargin=*]
    \item \textbf{state} $\boldsymbol{s}_t$: The state vector $\boldsymbol{s}_t$ encompasses a collection of features that characterizes the bidding conditions at timestep $t$. For advertising scenarios, these features could be the remaining time, unused budget, historical bidding statistics, etc.
    
    \item \textbf{action} $a_t$: The action $a_t$ denotes the bidding variables that could be iteratively adjusted through the whole bidding period. In this paper, according to Equation~\eqref{equ:bi}, the optimal action is $a = \lambda^*$. Therefore, we denote the real action at time step $t$ as:
    \begin{equation}\label{equ:al}
        a_t = \lambda_t
    \end{equation}
    \item \textbf{reward} $rw_t$: Suppose there are $N_t$ candidate impressions coming between $t$ and $t+1$. The reward $rw_t$ could then be defined as:

    \begin{equation}
    rw_t = \sum_{n=0}^{N_t} x_{n_t} v_{n_t}
    \end{equation}
    where $x_{n_t}$ and $v_{n_t}$ are the binary indicator and value of the $n^{th}$ impression at time step $t$.

    \item \textbf{Return-To-Go (RTG)} $r_t$: The RTG value indicates the total amount of rewards to be obtained in the future time steps:

    \begin{equation}\label{equ:problem2}
    r_t=\sum_{t^{\prime}=t}^T rw_{t^{\prime}}
    \end{equation}
    where $T$ is the final time step.
\end{itemize}

These settings result in the following trajectory representation, which is well-suited for autoregressive training and inference:

\begin{equation}
\tau=\left(r_1, \boldsymbol{s}_1, a_1, r_2, \boldsymbol{s}_2, a_2, \ldots, r_T, \boldsymbol{s}_T, a_T\right)
\end{equation}

\section{Framework}

Here, we will detail the \name's overview and key components.

\begin{figure*}[t]
    \centering
    \includegraphics[width=0.87\linewidth]{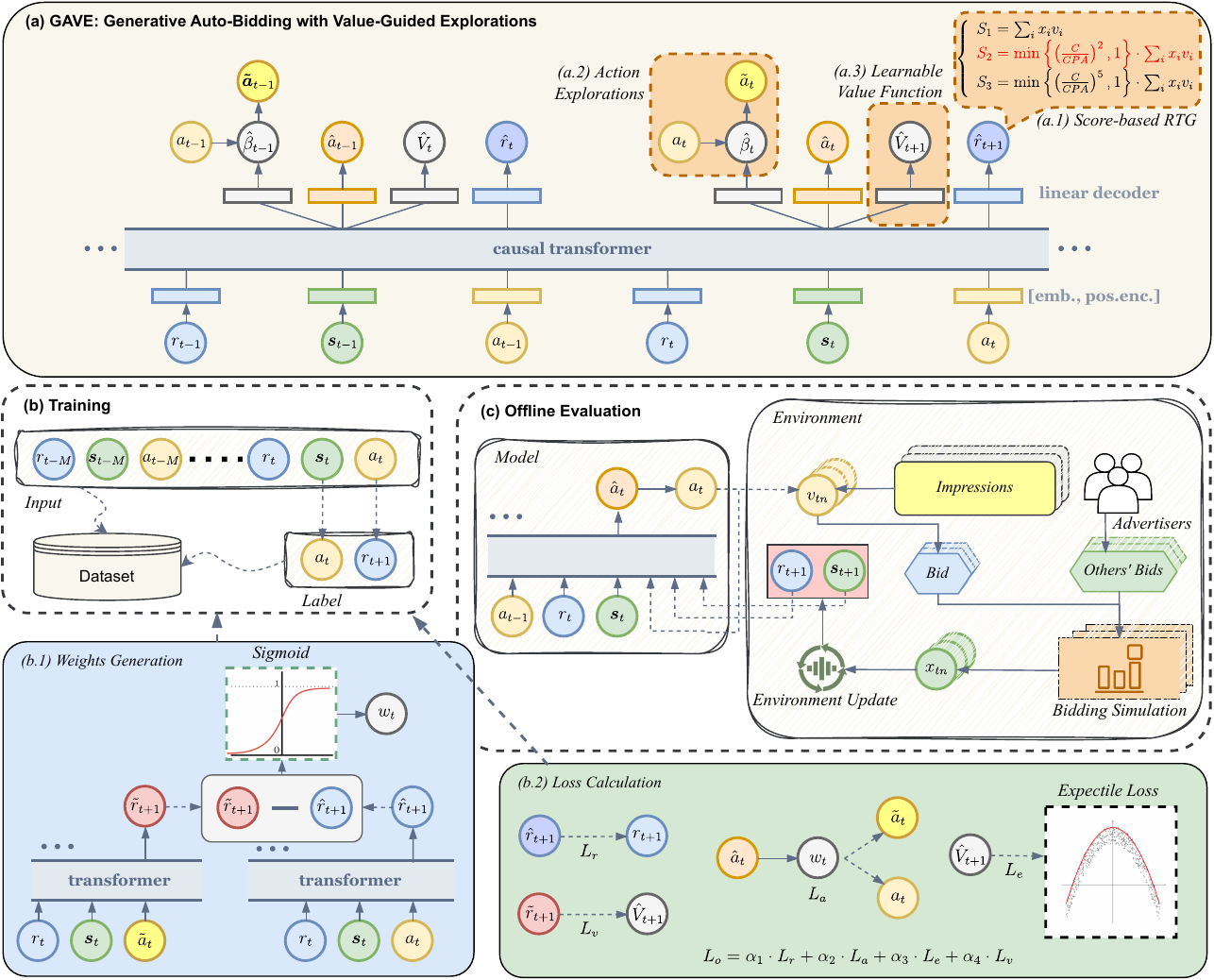}
    \vspace{-3mm}
    \caption{Overall structure of \name.}
    \label{fig:overview}
    \vspace{-5mm}
\end{figure*}

\subsection{GAVE Overview}

As shown in Figure~\ref{fig:overview}, \name adopts a DT architecture, where the pair of RTGs, states, and actions form the input sequence, i.e., $(r_t,\boldsymbol{s}_t,a_t)$ for the timestamp $t$.  Unlike conventional DTs, \name introduces several key innovations to achieve adaptive optimization, enhance stability, and facilitate strategy improvement. These include a \textbf{score-based RTG} (Figure~\ref{fig:overview} (a.1)) for aligning with diverse advertising objectives%evaluation metrics
, an \textbf{action exploration module} (Figure~\ref{fig:overview} (a.2)) equipped with an RTG-based evaluation mechanism for discovering and evaluating new actions and stabilizing updates, and a \textbf{learnable value function} (Figure~\ref{fig:overview} (a.3)) to steer exploration for strategy improvement while mitigating Out-of-Distribution (OOD) risks. The training of \name follows an offline paradigm (Figure~\ref{fig:overview} (b)), using sequence samples as input to generate predicted labels. For evaluation, a simulated bidding environment (Figure~\ref{fig:overview} (c)) is employed, where the test model interacts with fixed-policy agents.

Specifically, \name employs an adaptive score-based RTG function that can align optimization objectives with varying advertising objectives. During action exploration at time step $t$, in addition to predicting the action $\hat{a}_t$, \name predicts a coefficient $\hat{\beta}_t$, a $\hat{V}_{t+1}$ for estimating the learnable value function $V_{t+1}$, and an RTG value $\hat{r}_{t+1}$. The process is formally expressed as follows:

\begin{equation}\label{equ:predout}
\left\{
\begin{array}{l}
    (\hat{\beta}_t, \hat{a}_t, \hat{V}_{t+1})  = GAVE(r_{t-M}, \boldsymbol{s}_{t-M}, a_{t-M}, 
      \ldots, r_t, \boldsymbol{s}_t)\\
    \hat{r}_{t+1}  = GAVE(r_{t-M}, \boldsymbol{s}_{t-M}, a_{t-M},
      \ldots, r_t, \boldsymbol{s}_t, a_t)\\
    \tilde{a}_{t}  =  \hat{\beta}_t a_{t}
\end{array}
\right.
\end{equation}
where $M$ is a hyper-parameter, indicating a sequence with $M+1$ input time steps.

By evaluating the explored action $\tilde{a}_t$ and the action label $a_t$ with an RTG-based evaluation method, \name applies a balanced update strategy to reconcile $\tilde{a}_t$ and $a_t$. This ensures a stability-preserving update process. Additionally, the learnable value function $V_{t+1}$ is introduced to direct the model toward potentially optimal strategies while further reducing OOD risks. These innovations collectively enable \name to achieve improved performance and robustness.

\subsection{Score-based RTG}\label{sec:opt}

As illustrated in Section~\ref{sec:problem}, directly optimizing the cumulative value of won impressions may cause the CPA constraint to significantly exceed its limit. To address this issue, objective functions incorporating penalty terms can be established as evaluation metrics, allowing the degree of emphasis on CPA restrictions to be tailored to specific advertising objectives. This approach facilitates the evaluation and selection of the optimal model. For example, previous work~\cite{su2024a} proposes using a score $S$ to assess the model's actual performance during the testing phase, thereby enabling the selection of higher-performing models. This score integrates a penalty term for CPA constraints to evaluate the overall performance of the bidding model throughout the entire bidding period with $\gamma=2$:

\begin{equation}\label{equ:score}
\left\{
\begin{array}{l}
    CPA = \frac{\sum_i x_i c_i}{\sum_i x_i v_i} \\
    \mathbb{P}(CPA; C) = \min \left\{\left(\frac{C}{CPA}\right)^\gamma, 1\right\} \\
    S = \mathbb{P}(CPA; C) \cdot \sum_i x_i v_i 
\end{array}
\right.
\end{equation}

In this paper, we integrate the constraints directly into the training stage, moving beyond reliance on pre-trained model selection to achieve improved evaluation scores. To align with evaluation metrics for various advertising objectives, we propose employing a constrained score function instead of the unconstrained $\sum_{i=1}^I x_i v_i$ for RTG modeling in \name, as illustrated in Figure~\ref{fig:overview} (a.1). For example, based on the evaluation metric defined in Equation~\eqref{equ:score}, the following score-based RTG function can be utilized to synchronize training with evaluation:

\begin{equation}\label{equ:st}
    \left\{
    \begin{array}{l}
        CPA_t = \frac{\sum_i^{I_t} x_i c_i}{\sum_i^{I_t} x_i v_i} \\
        \mathbb{P}(CPA_t; C) = \min \left\{\left(\frac{C}{CPA_t}\right)^\gamma, 1\right\} \\
        S_t = \mathbb{P}(CPA_t; C) \cdot \sum_i^{I_t} x_i v_i \\
        r_t = S_T - S_{t-1}
    \end{array}
    \right.
\end{equation}

Here, $I_t$ denotes the number of impressions from time step $0$ to time step $t$, $S_t$ represents the generalized score function at time step $t$, and $T$ signifies the final time step in a bidding period. By generalizing the score calculation to each time step, the RTG $r_t$ is derived to represent the future score yet to be obtained, guiding the optimization direction of \name.

Furthermore, in practical applications, different advertising objectives may exhibit varying degrees of dependence on CPA constraints, resulting in different evaluation metrics. Nonetheless, training and evaluation can remain aligned by generalizing $S$ to each time step (i.e., $S_t$) in a similar way, leading to:

\begin{equation}\label{equ:genr}
    r_t = S_T - S_{t-1}
\end{equation}

This score-based RTG function enhances the flexibility of \name, ensuring its applicability across diverse advertising objectives.

\subsection{Action Explorations}\label{sec:moduleaction}

The primary objective of this section is to explore novel actions during training to discover strategies potentially absent from the offline dataset, thereby enabling better model optimization. However, in an offline setting where environment interaction is impossible,
learning merely from the fixed dataset may lead to documented behavioral patterns. Conversely, exploring actions beyond the dataset can introduce inherent distribution shifts, potentially leading to behavioral collapse~\cite{chen2021decision,kostrikovoffline}.
% feedback values for explored actions must be estimated rather than directly observed. This limitation introduces inherent distribution shifts when utilizing explored actions for training. 
Moreover, compared to real action labels, the impact of explored actions on model performance can be either beneficial or detrimental, presenting significant challenges in developing a stability-preserving update procedure.

To address these challenges, \name introduces a novel action exploration mechanism in conjunction with an RTG-based evaluation method as illustrated in Figure~\ref{fig:overview} (a.2). This enables \name to adaptively adjust both the exploration and update directions of actions by identifying their significance, thereby achieving stability-preserving updates. Specifically, at time step $t$, \name predicts a coefficient $\hat{\beta}_t$ with the same dimensionality as $a_t$ to generate a new action $\tilde{a}_t$. This process is formally expressed as:

\begin{equation}
\left\{
\begin{array}{l}
    \hat{\beta}_t = \sigma(FC_{\beta}(DT(r_{t-M}, \boldsymbol{s}_{t-M}, a_{t-M}, 
      \ldots, r_t, \boldsymbol{s}_t)))\\
      \tilde{a}_{t}  =  \hat{\beta}_t a_{t}
\end{array}
\right.
\end{equation}
where $DT()$ represents the DT backbone, $FC_{\beta}()$ denotes a fully-connected layer, and $\sigma$ is scaling function. To mitigate OOD issues, the scaling function is defined as:

\begin{equation}
    \sigma(x) = Sigmoid(x)+0.5
\end{equation}

This formulation constrains $\hat{\beta}_t$ to the interval $(0.5, 1.5)$, ensuring the explored action $\tilde{a}_t$ remains in proximity to the action label $a_t$.

To minimize distribution shift and obtain stability-preserving updates during training, rather than directly utilizing $\tilde{a}_t$ for generating new samples, we employ it as an additional label to balance action updates in conjunction with the original label $a_t$. This approach necessitates estimating the relative significance of $\tilde{a}_t$ and $a_t$ to determine an optimal update direction for the predicted action $\hat{a}_t$. Following reinforcement learning conventions~\cite{mnih2013playing,nachum2017bridging,kostrikovoffline}, we define the action-value of $a_t$ as $r_{t+1}$ (the RTG at time step $t+1$), as it represents the cumulative future returns after executing action $a_t$. This enables the design of $w_t$ as illustrated in Figure~\ref{fig:overview} (b.1) to balance the update direction:

\begin{equation}\label{equ:predr}
\left\{
    \begin{array}{l}
    \tilde{r}_{t+1} = \name(r_{t-M}, \boldsymbol{s}_{t-M}, a_{t-M}, 
      \ldots, r_t, \boldsymbol{s}_t, \tilde{a}_t)))\\
    w_t = Sigmoid(\alpha_r \cdot (\tilde{r}_{t+1} - \hat{r}_{t+1}))
    \end{array}
\right.
\end{equation}
where $\tilde{r}_{t+1}$ and $\hat{r}_{t+1}$ represent the estimated RTG for $\tilde{a}_t$ and $a_t$ respectively. The corresponding loss function for action explorations is defined as:

\begin{equation}\label{equ:lra}
\left\{
\begin{aligned}
    L_r &= \frac{1}{M+1} \sum_{t-M}^{t}{(\hat{r}_{t+1} - r_{t+1})^2}\\
    L_a &= \frac{1}{M+1} \sum_{t-M}^{t}{((1-w^\prime_t) \cdot (\hat{a}_t - a_t)^2 + w^\prime_t \cdot (\hat{a}_t - \tilde{a}^\prime_t)^2)}
\end{aligned}
\right.
\end{equation}
where $w^\prime$ and $\tilde{a}^\prime_t$ denote $w$ and $\tilde{a}_t$ with frozen gradients. Through $L_r$, \name ensures accurate RTG prediction, enabling reliable estimation of the RTG for both $\tilde{a}_t$ and $a_t$. Through $L_a$, \name maintains a balanced and stability-preserving updating process between $\tilde{a}_t$ and $a_t$, directing updates toward $\tilde{a}_t$ when it proves superior ($w_t>0.5$), and toward $a_t$ otherwise to mitigate OOD issues and potential negative impacts from exploration.

\subsection{Learnable Value Function}\label{sec:modulevalue}

While the action exploration mechanism ensures explorations outside the dataset and a stability-preserving update process, randomly generated $\tilde{a}_t$ cannot guarantee improved model performance. To address this limitation, we propose a learnable value function that facilitates the discovery of superior actions for strategy improvement, as illustrated in Figure~\ref{fig:overview} (a.3). Specifically, drawing inspiration from reinforcement learning conventions~\cite{mnih2013playing,nachum2017bridging,kostrikovoffline}, we propose a sequence-value function $V_{t+1}$ analogous to the optimal state-value function in RL, which represents the upper bound of $r_{t+1}$ as follows:

\begin{equation}\label{equ:value}
    V_{t+1} = \mathop{\arg\max}_{a_t \in \mathbb{A}} {r_{t+1}}
\end{equation}
where $\mathbb{A}$ denotes the available action space. Due to the extensive action space and limited real actions within offline datasets, the direct statistical computation of $V_{t+1}$ is infeasible. However, we can learn this value through an expectile regression process with $r_{t+1}$:

\begin{equation}\label{equ:le}
    \begin{aligned}
    L_e &= \frac{1}{M+1} \sum_{t-M}^{t}{(L_2^\tau(r_{t+1}-\hat{V}_{t+1}))}\\
    &= \frac{1}{M+1} \sum_{t-M}^{t}{(|\tau - \mathbbm{1}((r_{t+1}-\hat{V}_{t+1}) < 0)| (r_{t+1}-\hat{V}_{t+1})^2)}
    \end{aligned}
\end{equation}
where $L_2^\tau\left(y-m(x)\right)$ represents the loss function for predicting the expectile $\tau \in (0, 1)$ of labels $y$ with model $m(x)$~\cite{kostrikovoffline}. $\hat{V}_{t+1}$ represents the predicted value of $V_{t+1}$. Following Equation~\eqref{equ:value}, we set $\tau=0.99$ to learn the upper bound of $r_{t+1}$, effectively estimating $V_{t+1}$. 

\begin{algorithm}[t]
	\caption{\label{alg:GAVE} Optimization algorithm of \name}
	\label{alg:1}
	\raggedright
	{\bf Input}: A training dataset $\mathcal{D}=\left\{\left(\boldsymbol{z}_j, \boldsymbol{y}_j\right)\right\}_{j=1}^{|\mathcal{D}|}$ with $|\mathcal{D}|$ samples sampled from a bidding environment. $\boldsymbol{z}_j$ is a sequence with $M+1$ time steps and $\boldsymbol{y}_j$ are the label set. $\boldsymbol{z}_j=\{r_{t-M}, \boldsymbol{s}_{t-M}, a_{t-M}, 
      \ldots, r_t, \boldsymbol{s}_t, a_t \}$; $\boldsymbol{y}_j =\{a_t, r_{t+1}\}$\\
	{\bf Output}: A well-trained model $f$ with parameters $\boldsymbol{\Phi}$\\

	\begin{algorithmic} [1]
	    \STATE Randomly initialize parameters $\boldsymbol{\Phi}$ of the model $f$
        \FOR{Step 1,..., Max Step}
            \STATE Sample a training batch $B$ from $\mathcal{D}$
            \STATE Obtain $\hat{\beta}_t$, $\hat{a}_t$, $\hat{V}_{t+1}$ , $\hat{r}_{t+1}$ and $\tilde{a}_t$ with $f(B)$ via Equation~\eqref{equ:predout}
            \STATE Obtain $\tilde{r}_{t+1}$ and $w_t$ via Equation~\eqref{equ:predr}
            \STATE Calculate loss $L_r$, $L_a$, $L_e$ and $L_v$ based on Equation~\eqref{equ:lra}, ~\eqref{equ:le} and ~\eqref{equ:lv}
            \STATE Calculate the overall loss $L_o$ based on Equation~\eqref{equ:lo}
            \STATE Update $\boldsymbol{\Phi}$ via minimizing the loss $L_o$
        \ENDFOR
        \STATE return $f$~\\
        
	\end{algorithmic}
\end{algorithm}

By estimating $V_{t+1}$ with $\hat{V}_{t+1}$ and utilizing it to guide the updating directions of $\tilde{r}_{t+1}$, \name implicitly steers the update direction of the explored $\tilde{a}_t$ toward potentially optimal actions. This process is illustrated in Figure~\ref{fig:overview} (b.2) and can be formalized as:

\begin{equation}~\label{equ:lv}
    L_v = \frac{1}{M+1} \sum_{t-M}^{t}{(\tilde{r}_{t+1}-\hat{V}^\prime_{t+1})^2}  
\end{equation}
where $\hat{V}^\prime_{t+1}$ represents the gradient-frozen version of $\hat{V}_{t+1}$. Through the application of $L_v$, 
% \name ensures that $\tilde{r}_{t+1}$ updates toward the potential optimal RTG, thereby implicitly guiding the update direction of $\tilde{a}_{t}$ toward the optimal action at time step $t$.
\name implicitly guides the update direction of $\tilde{a}_{t}$ toward the optimal action by anchoring their RTGs near $\hat{V}_{t+1}$. This approach mitigates OOD risks and enables controlled extrapolation for strategy improvement.

% \begin{figure}[t]

% \vspace{-12mm}
% \end{figure}

% \subsection{Optimization Algorithm}

% Through the above designs, \name realizes an offline generative auto-bidding framework with value-guided explorations for better strategy learning. The final loss function could be expressed as a weighted sum of Equation~\eqref{equ:lra}, ~\eqref{equ:le} and ~\eqref{equ:lv}, which is:

% \begin{equation}\label{equ:lo}
%     L_o = \alpha_1 \cdot L_r + \alpha_2 \cdot L_a + \alpha_3 \cdot L_e + \alpha_4 \cdot L_v  
% \end{equation}
% where $\alpha_1$, $\alpha_2$, $\alpha_3$ and $\alpha_4$ are hyper-parameters. The pseudocode of \name's overall optimization algorithm is shown in Algorithm~\ref{alg:GAVE}. The training process is also illustrated in Figure~\ref{fig:overview} (b).

% In the inference stage, as illustrated in Figure~\ref{fig:overview} (c), for each input sequence, \name predicts the corresponding $\hat{a}_t = \lambda_t$ as the bid parameters at time step $t$, and the bidding price for the n-th impression at $t$ could be obtained via $b_{tn} = \lambda_t v_{tn}$ as illustrated in Equation~\eqref{equ:bi} for bidding simulation.
\subsection{Optimization Algorithm}

Through the mechanisms described above, \name implements an offline generative auto-bidding framework incorporating value-guided explorations to enhance strategy learning. The comprehensive loss function is formulated as a weighted combination of the components defined in Equations~\eqref{equ:lra}, ~\eqref{equ:le}, and ~\eqref{equ:lv}:

\begin{equation}\label{equ:lo}
    L_o = \alpha_1 \cdot L_r + \alpha_2 \cdot L_a + \alpha_3 \cdot L_e + \alpha_4 \cdot L_v  
\end{equation}
where $\{\alpha_1, \alpha_2, \alpha_3, \alpha_4\}$ are hyperparameters controlling the relative contribution of each loss component. The complete optimization procedure of \name is detailed in Algorithm~\ref{alg:GAVE}, with the training process visualized in Figure~\ref{fig:overview} (b).

During inference, as illustrated in Figure~\ref{fig:overview} (c), \name processes each input sequence to predict $\hat{a}_t = \lambda_t$, which serves as the bid parameter at time step $t$. The bidding price for the $n$-th impression at time step $t$ is then computed according to Equation~\eqref{equ:bi} as $b_{tn} = \lambda_t v_{tn}$, enabling real-time bidding simulation.

\section{Offline Experiments}
In this section, we conduct experiments on two public datasets to investigate the following questions:

\begin{itemize}[leftmargin=*]
    \item \textbf{RQ1:} How does \name perform compared to state-of-the-art auto-bidding baselines?
    \item \textbf{RQ2:} Can \name adapt to diverse advertising objectives?
    \item \textbf{RQ3:} How effective is the proposed learnable value function in facilitating action exploration?
    \item \textbf{RQ4:} How do the proposed components in \name contribute to the final bidding performance?
\end{itemize}

In the following subsections, we begin by outlining the evaluation settings. Then, we address the relevant questions by concisely analyzing our experimental findings.

% Table generated by Excel2LaTeX from sheet 'Sheet2'
\begin{table}[t]
  \centering
  \caption{Data statistics.}
  \vspace{-4mm}
  \scalebox{0.7}{
    \begin{tabular}{ccc}
    \toprule
    Params & AuctionNet & AuctionNet-Sparse \\
    \midrule
    Trajectories & 479,376 & 479,376 \\
    Delivery Periods & 9,987 & 9,987 \\
    Time steps in a trajectory & 48    & 48 \\
    State dimension & 16    & 16 \\
    Action dimension & 1     & 1 \\
    Return-To-Go Dimension & 1     & 1 \\
    Action range & [0, 493] & [0, 589] \\
    Impression's value range & [0, 1] & [0, 1] \\
    CPA range & [6, 12] & [60, 130] \\
    Total conversion range & [0, 1512] & [0, 57] \\
    \bottomrule
    \end{tabular}%
    }
  \label{tab:statistics}%
  \vspace{-5mm}
\end{table}%

\subsection{Experimental Setup}

% Table generated by Excel2LaTeX from sheet 'Sheet2'
\begin{table*}[t]
  \centering
  \caption{Performance comparison. The boldface denotes the highest score. The underline indicates the best result of baselines. ``\textbf{{\Large *}}'' indicates the statistically significant improvements (i.e., two-sided t-test with $p<0.05$) over the best baseline.}
  \vspace{-3mm}
  \setlength\tabcolsep{9pt}
  \scalebox{0.8}{
    \begin{tabular}{c|c|cccccccccc}
    \toprule
    \toprule
    Dataset & Budget & DiffBid & USCB  & CQL   & IQL   & BCQ   & DT    & CDT   & GAS   & GAVE  & \textit{Improve} \\
    \midrule
    \multirow{5}[2]{*}{AuctionNet} & 50\%  & 54    & 86    & 113   & 164   & 190   & 191   & 174   & \underline{193}   & \textbf{201*} & 4.15\% \\
          & 75\%  & 100   & 135   & 139   & 232   & 259   & 265   & 242   & \underline{287}   & \textbf{296*} & 3.14\% \\
          & 100\% & 152   & 157   & 171   & 281   & 321   & 329   & 326   & \underline{359}   & \textbf{376*} & 4.74\% \\
          & 125\% & 193   & 220   & 201   & 355   & 379   & 396   & 378   & \underline{409}   & \textbf{421*} & 2.93\% \\
          & 150\% & 234   & 281   & 238   & 401   & 429   & 450   & 433   & \underline{461}   & \textbf{467*} & 1.30\% \\
    \midrule
    \multirow{5}[2]{*}{AuctionNet-Sparse} & 50\%  & 9.9   & 11.5  & 12.8  & 16.5  & 17.7  & 14.8  & 11.2  & \underline{18.4}  & \textbf{19.6*} & 6.52\% \\
          & 75\%  & 15.4  & 14.9  & 16.7  & 22.1  & 24.6  & 22.9  & 18.0  & \underline{27.5}  & \textbf{28.3*} & 2.91\% \\
          & 100\% & 19.5  & 17.5  & 22.2  & 30.0  & 31.1  & 29.6  & 31.2  & \underline{36.1}  & \textbf{37.2*} & 3.05\% \\
          & 125\% & 25.3  & 26.7  & 28.6  & 37.1  & 34.2  & 34.3  & 31.7  & \underline{40.0}  & \textbf{42.7*} & 6.75\% \\
          & 150\% & 30.8  & 31.3  & 35.8  & 43.1  & 37.9  & 44.5  & 39.1  & \underline{46.5}  & \textbf{47.4*} & 1.94\% \\
    \bottomrule
    \bottomrule
    \end{tabular}%
    }
  \label{tab:overall}%
  \vspace{-4mm}
\end{table*}%

\subsubsection{\textbf{Dataset}}
Previous auto-bidding research has predominantly relied on proprietary bidding logs for evaluation, with problem formulations often specific to particular scenarios. This heterogeneity in evaluation methodologies has hindered fair and systematic comparisons across different approaches. Recently, Alimama introduced AuctionNet~\footnote{https://github.com/alimama-tech/AuctionNet}~\cite{su2024a}, the industry's first standardized large-scale simulated bidding benchmark, enabling comprehensive model evaluation under consistent conditions.

In this study, we utilize two datasets from the AuctionNet framework: (i) \textbf{AuctionNet}: The primary dataset containing comprehensive bidding trajectories, and (ii) \textbf{AuctionNet-Sparse}: A sparse variant of AuctionNet featuring reduced conversion rates. Both datasets comprise approximately 500k bidding trajectories collected across 10k distinct delivery periods, each consisting of 48 time steps and interactions derived from millions of impression opportunities. Detailed statistics are presented in Table~\ref{tab:statistics}.

\subsubsection{\textbf{Evaluation Protocol}}
Our evaluation methodology follows the AuctionNet benchmark~\cite{su2024a} and employs a simulated environment~\cite{li2024gas} that emulates real-world advertising systems, as illustrated in Figure~\ref{fig:overview} (c). The evaluation spans a 24-hour delivery period, discretized into 48 uniform time steps, during which the predicted action is utilized for bidding ($\hat{a}_t = a_t$). Within this simulated environment, 48 bidding agents with distinct strategies compete for incoming impression opportunities, with performance measured using Equation~\eqref{equ:score} with $\gamma=2$.

To ensure comprehensive evaluation, we employ a round-robin testing strategy: the test model sequentially replaces each of the 48 agents, competing against the remaining agents in each round. The final performance is computed as the average score across all evaluations, providing a robust measure of the model's effectiveness.

\subsubsection{\textbf{Baselines}}

To assess the effectiveness of \name, we perform a thorough comparison with multiple baseline approaches:

\begin{itemize}[leftmargin=*]
    \item \textbf{DiffBid}~\cite{guo2024generative}: applies the diffusion frameworks to simulate bidding trajectories and model bidding sequences. 
    \item \textbf{USCB}~\cite{he2021unified}: dynamically adjust bidding parameters for optimal bidding performance in an online RL bidding environment.
    \item \textbf{CQL}~\cite{kumar2020conservative}: learns a conservative value function to mitigate the overestimation problems in offline RL.
    \item \textbf{IQL}~\cite{kostrikovoffline}: applies an expectile regression method to enable policy improvement without evaluating the out-of-scope actions.
    \item \textbf{BCQ}~\cite{fujimoto2019off}: applies a restriction on the action space for a typical offline RL learning process.
    \item \textbf{DT}~\cite{chen2021decision}: employs a transformer architecture for sequential decision-making modeling and utilizes a behavior cloning method to learn the average strategy from the dataset.
    \item \textbf{CDT}~\cite{liu2023constrained}: tries to train a constraint satisfaction policy in the offline settings for a balance of safety and task performance.
    \item \textbf{GAS}~\cite{li2024gas}: tries to model a DT-based offline bidding framework with post-training search by applying Monte Carlo Tree Search (MCTS) in modeling. 
\end{itemize}

\subsubsection{\textbf{Implementation Details}}
Following previous studies~\cite{li2024gas, su2024a}, we conduct evaluations using varying budget ratios from the original dataset. Performance is measured using the scoring metric:

\begin{equation}\label{equ:score1}
    S=\mathbb{P}(C P A ; C) \cdot \sum_i x_i  v_i 
\end{equation}
as defined in Equation~\eqref{equ:score} with $\gamma=2$.

All experiments are conducted on NVIDIA H100 GPUs, utilizing a fixed batch size 128 for a maximum of 400k training steps. The \name implementation employs a causal transformer architecture with 8 layers and 16 attention heads. Model parameters are optimized using the AdamW optimizer~\cite{loshchilov2017decoupled} with a learning rate of $1e^{-5}$. Additional hyper-parameters are determined through a comprehensive grid search to maximize performance. To ensure statistical significance, we conduct 10 independent runs using the optimal parameter configuration and report the average performance metrics.

\subsection{Overall Performance (RQ1)}

We present a comprehensive comparison between \name and various baseline approaches across different budget settings, with results summarized in Table~\ref{tab:overall}. Our experimental analysis reveals several key findings:

\begin{itemize}[leftmargin=*]
    \item \name demonstrates superior performance across all budget and dataset configurations, consistently outperforming existing methods. This superiority can be attributed to our novel action exploration method with the value function's guidance, which enables the discovery of novel, potentially optimal actions beyond the offline dataset while maintaining robust training through a stability-preserving update process balancing the exploration benefits and risks.
    \item Among all baselines, DT-based methods (GAS, DT, and CDT) exhibit superior performance, highlighting the effectiveness of DT structure in capturing temporal dependencies and facilitating sequential decision-making in bidding scenarios. Notably, GAS achieves better results compared to DT and CDT, validating the efficacy of its MCTS implementation in strategy optimization.
    \item DiffBid does not perform well on the datasets, probably due to the long sequence and highly dynamic environment posing extra challenges for DiffBid in accurately predicting trajectories and learning from its reverse process. 
\end{itemize}

\subsection{Alignment Analysis (RQ2)}

As discussed in Section~\ref{sec:opt}, advertising objectives may necessitate different evaluation metrics. To address this, \name employs an adaptive score-based RTG modeling that accommodates various optimization objectives, thereby aligning training objectives with evaluation metrics as illustrated in Equation~\eqref{equ:genr}. In this section, we explore the performance of \name under different RTG and evaluation metric configurations to answer RQ2. Specifically, we consider three evaluation metrics defined as follows:

\begin{equation}\label{eq:metrics}
    \left\{
    \begin{array}{l}
        S_1 = \sum_i x_i v_i, \\
        S_2 = \min \left\{\left(\frac{C}{CPA}\right)^2, 1\right\} \cdot \sum_i x_i v_i, \\
        S_3 = \min \left\{\left(\frac{C}{CPA}\right)^5, 1\right\} \cdot \sum_i x_i v_i,
    \end{array}
    \right.
\end{equation}
where $S_1$ only considers the total impression values obtained and represents the business scenarios with elastic restrictions on CPA conditions. $S_2$ is the optimization goal and evaluation score of this paper. It adds a punishment on CPA constraints. $S_3$ further enhances the penalty coefficient of CPA to represent the business scenarios with strict restrictions on CPA conditions. These metrics are utilized either for RTG modeling during training or as evaluation criteria during testing. The results are presented in Table~\ref{tab:align}.

\begin{table}[t]
  \centering
  \caption{Alignment analysis on AuctionNet-Sparse with 100\% budget. ``Train'' denotes applying the score to RTG modeling, while ``Eval'' denotes its utilization as the evaluation metric.}
  \vspace{-3mm}
    \setlength\tabcolsep{20pt}
    \scalebox{0.8}{
    \begin{tabular}{c|ccc}
    \toprule
    Train \textbackslash{} Eval & $S_1$    & $S_2$    & $S_3$ \\
    \midrule
    $S_1$    & \cellcolor[HTML]{C5D5E7}41.4  & \cellcolor[HTML]{F3F3F9}33.0  & \cellcolor[HTML]{F3F3F9}23.6  \\
    $S_2$    & \cellcolor[HTML]{D7E6EE}39.9  & \cellcolor[HTML]{C5D5E7}37.2  & \cellcolor[HTML]{D7E6EE}33.3  \\
    $S_3$    & \cellcolor[HTML]{F3F3F9}39.1  & \cellcolor[HTML]{D7E6EE}36.8  & \cellcolor[HTML]{C5D5E7}33.5  \\
    \bottomrule
    \end{tabular}%
    }
  \label{tab:align}%
  \vspace{-4mm}
\end{table}%

From Table~\ref{tab:align}, we observe that \name consistently achieves the highest performance when the training RTG corresponds to the same function used as the evaluation metric. This finding underscores the importance of aligning training objectives with specific evaluation metrics through our score-based RTG approach.

% \subsection{Parameter Analysis (RQ3)}
% To answer RQ3, in this section, a parameter analysis is conducted on the weight $w_t$ as shown in Figure~\ref{fig:overview} (b.1) to illustrate the difference of $\tilde{a}_t$ and $a_t$ in the training process. Specifically, we visualize the average $L_o$ and $w_t$ at training steps in Figure~\ref{fig:hyper} to track the difference between $\tilde{r}_t$ and $\hat{r}_t$. The larger $w_t$ is, the higher $\tilde{r}_t$ is over $\hat{r}_t$, thus proving that $\tilde{a}_t$ is better than $a_t$, indicating the effectiveness of the value function's guidance on action explorations.

% From Figure~\ref{fig:hyper}, we observe that as the training progresses and the overall loss $L_o$ decreases, the parameter $w$ increases from around 0.5 to a stable position above 0.5, verifying the effectiveness of the value function's guidance on action explorations. The stable position is influenced by both the dataset distribution and the model's hyper-parameters.  With the guidance of the value function $V$, the model can continuously explore actions with larger RTG $r$ in a plausible range, facilitating learning from the potential optimal strategy while alleviating OOD issues.

% \begin{figure}[t]
%     \centering
%     \includegraphics[width=1.0\linewidth]{fig/w.pdf}
%     \caption{Parameter analysis of $w$ on AuctionNet.}
%     \label{fig:hyper}
% \end{figure}
\subsection{Parameter Analysis (RQ3)}

To address RQ3, we conduct a parameter analysis of the weight $w_t$, as depicted in Figure~\ref{fig:overview} (b.1), to elucidate the distinction between $\tilde{a}_t$ and $a_t$ during the training process. Specifically, Figure~\ref{fig:hyper} visualizes the average overall loss $L_o$ and the weight $w_t$ across training steps, allowing us to monitor the disparity between $\tilde{r}_{t+1}$ and $\hat{r}_{t+1}$. A larger $w_t$ signifies a greater influence of $\tilde{r}_{t+1}$ over $\hat{r}_{t+1}$, thereby demonstrating that $\tilde{a}_t$ is superior to $a_t$. This result underscores the effectiveness of the value function in guiding action exploration.

From Figure~\ref{fig:hyper}, it is evident that as training progresses, the parameter $w_t$ increases from approximately 0.5 to a stable position above 0.5. The stable position is influenced by both the dataset distribution and the model's hyper-parameters. This trend confirms the efficacy of the learnable value function in directing action exploration. With the guidance of the value function, the model consistently explores actions $\tilde{a}$ with higher RTG values $\tilde{r}_{t+1}$ near the estimated optimal value $\hat{V}_{t+1}$. This approach facilitates learning potential optimal strategies while mitigating OOD issues.

\begin{figure}[t]
    \centering
    \includegraphics[width=0.7\linewidth]{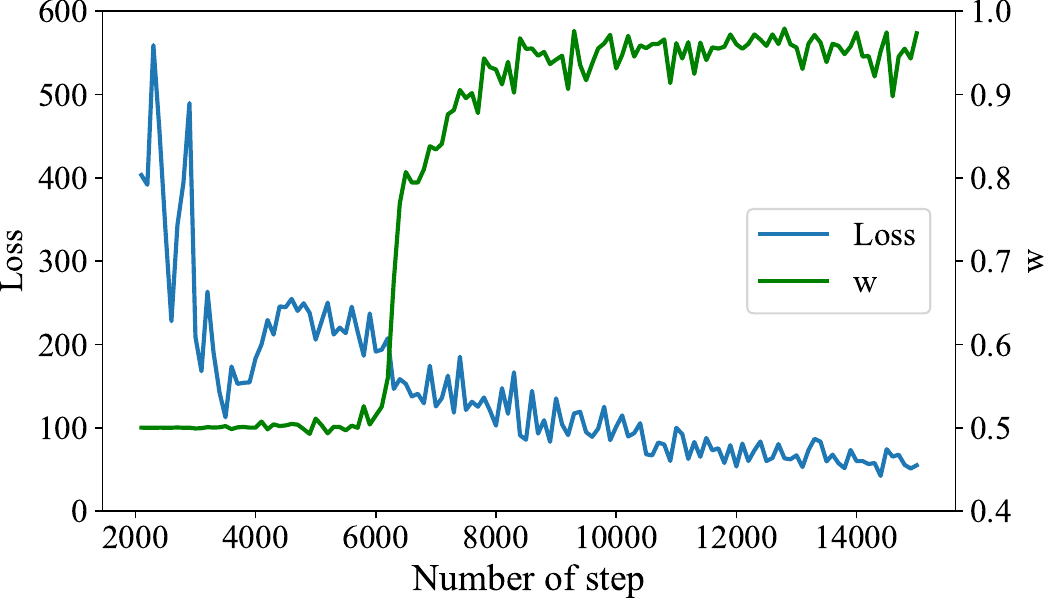}
    \vspace{-3mm}
    \caption{Parameter analysis of $w$ on AuctionNet.}
    \label{fig:hyper}
    \vspace{-5mm}
\end{figure}

\subsection{Ablation Study (RQ4)}

To further elucidate the contribution of each module within \name for answering RQ4, we conduct an ablation study by evaluating the following modified versions of \name:

\begin{itemize}[leftmargin=*]
    \item \textbf{\name-V}: excludes the learnable value function described in Section~\ref{sec:modulevalue}. In this configuration, the loss functions $L_v$ and $L_e$ are replaced with the following update rule to ensure that the explored actions generally surpass the original labels by enhancing their RTG values $\tilde{r}_{t+1}$:
    \begin{equation}
        L_w = 1 - \text{Sigmoid}(\alpha_r \cdot (\tilde{r}_{t+1} - \hat{r}^\prime_{t+1}))
    \end{equation}
    where $\hat{r}^\prime_{t+1}$ is a gradient-frozen version of $\hat{r}_{t+1}$. However, without the value function, the update direction of $\tilde{r}_{t+1}$ becomes unbounded, leading to OOD issues and suboptimal performance.
    
    \item \textbf{\name-VA}: omits both the value function from Section~\ref{sec:modulevalue} and the action exploration mechanism detailed in Section~\ref{sec:moduleaction}.
    
    \item \textbf{DT}: removes all designed modules related to \name, including those described in Sections~\ref{sec:modulevalue}, \ref{sec:moduleaction}, and Section~\ref{sec:opt}. Consequently, this configuration aligns with the pure DT framework~\cite{chen2021decision} with $S = \sum_i x_i v_i$ for RTG modeling.
\end{itemize}

\begin{figure}[t]
    \centering
    \subfigure[AuctionNet.]{
        \label{fig:ab1}
        \includegraphics[width=0.44\linewidth]{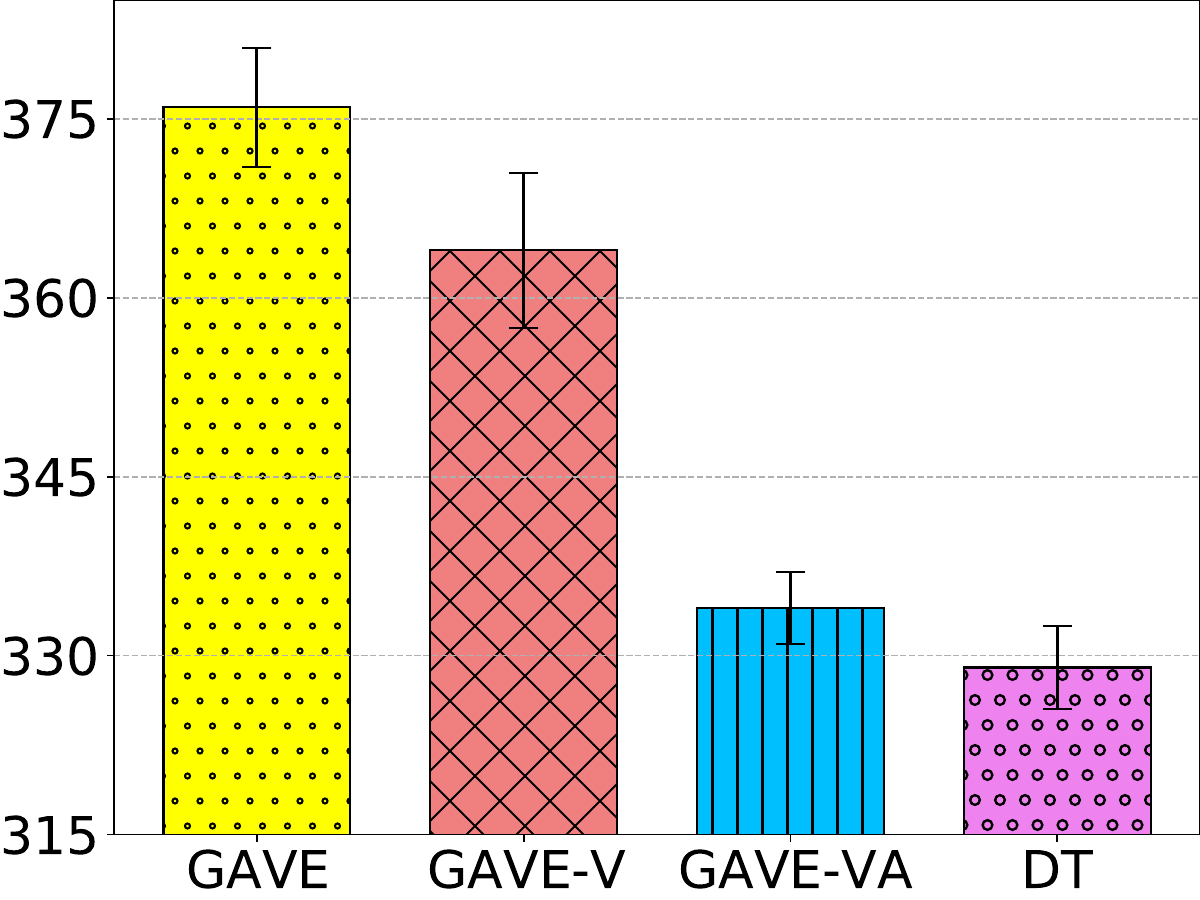}}
    \subfigure[AuctionNet-Sparse.]{
        \label{fig:ab2}
        \includegraphics[width=0.44\linewidth]{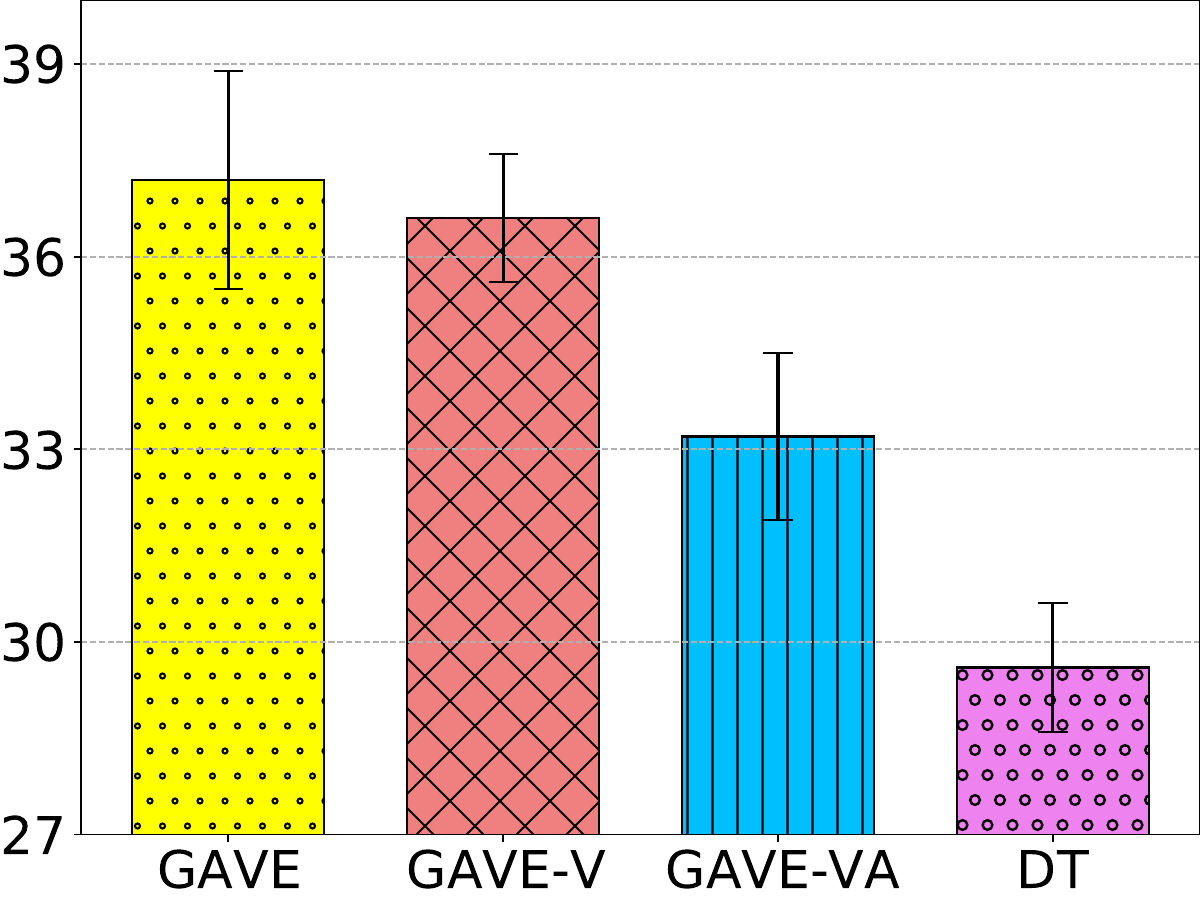}}
        \vspace{-5mm}
    \caption{Ablation study with 100\% budget.}
    \label{fig:ablation}
    \vspace{-5mm}
\end{figure}

Figure~\ref{fig:ablation} presents the evaluation results. The findings indicate that: (i) Aligning the optimization objectives with the evaluation metrics using the score-based RTG modeling allows \name-VA to outperform DT, demonstrating the importance of objective alignment in training; (ii) Incorporating the action exploration mechanism and RTG-based evaluation in \name-V enables the model to discover potential strategies beyond the offline dataset and evaluate their significance for a stability-preserving update process, thereby achieving better performance than \name-VA; and (iii) Fully integrating the value function to guide action explorations within \name leverages potential optimal strategies, further alleviating OOD issues and enhancing overall performance.

\section{Online Application}

We evaluate \name's effectiveness through A/B tests in two industrial live bidding scenarios: Nobid~\footnote{https://support.google.com/google-ads/answer/7381968?hl=en} (maximizing conversions within daily budget) and Costcap (maximizing conversions with CPA/ROI limits). The experimental settings are detailed below.

\begin{itemize}[leftmargin=*]
    \item \textbf{State}: 20-step sequence with features including budget, CPA limit, predictions, traffic/cost speeds, time-phased budget, remaining time, and window-averaged bid coefficient.
    
    \item \textbf{Action}: To stabilize bidding results, the bid coefficient $\lambda$ is determined based on a windowed-average of the preceding two hours containing $E$ time steps, $\lambda_t = a_t + \frac{1}{|E|}\sum_{t^\prime=t-E}^{t-1} \lambda_{t^\prime}$, where $a_t$ is \name's output action at time step $t$.
    
    \item \textbf{Return-To-Go (RTG)}: Given the sparsity of real conversions, we utilize the expected total conversions, $\sum_i pcvr_i$ during training, where $pcvr_i$ is the predicted conversion rate for winning traffic $i$. During inference, the RTG for the entire sequence is set to the campaign's total expected conversions from the previous day.

\end{itemize}

We compare \name with the offline reinforcement learning algorithm IQL~\cite{kostrikovoffline}, currently in production. Evaluation metrics include cost, conversions, target cost, and CPA valid ratio, with bidding strategies focused on maximizing conversions within budget and CPA constraints. To account for varying campaign targets, target cost serves as a value-weighted conversion measure. For Costcap campaigns, the conversion value equals the CPA limit, while Nobid campaigns use the average real CPA from total traffic. A Costcap campaign is CPA valid if its CPA remains below the limit, assessed solely for Costcap campaigns. Our five-day online A/B testing allocates 25\% of each campaign's budget and traffic to the baseline bidding model and \name, with results summarized in Table~\ref{tab:online}.

For both Nobid and Costcap, \name improves cost and conversions. In Nobid campaigns, \name achieves a 0.8\% increase in cost, 8.0\% in conversions, and 3.2\% in target cost. For Costcap campaigns, advertising revenue and advertiser value rise alongside a significant improvement in CPA validity, with +2.0\% cost, +3.6\% conversions, +2.2\% target cost, and +1.9\% valid CPA ratio.

\begin{table}[]
  \centering
  \caption{Online A/B test.}
  \vspace{-4mm}
  \scalebox{0.9}{
\begin{tabular}{l|cccc}
\hline
         & Cost & Conversion & Target cost & CPA valid ratio \\ \hline
Nobid    & +0.8\%      &  +8.0\%      &  +3.2\%   & /               \\ \hline
Costcap  & +2.0\%      & +3.6\%           &   +2.2\%       & +1.9\%           \\ \hline
\end{tabular}
}
\label{tab:online}
\vspace{-4mm}
\end{table}
\section{Related Works}

This section provides a brief review of relevant research topics, i.e., offline reinforcement learning, and auto-bidding.

\subsection{Offline Reinforcement Learning and Decision Transformers}  
Reinforcement Learning (RL) trains decision-making agents through environment interactions, evolving from foundational works~\cite{minsky1961steps,michie1968boxes} to advanced methods like policy gradient~\cite{sutton1999policy}, deep Q-learning~\cite{mnih2013playing}, and deterministic policy optimization~\cite{silver2014deterministic}. While effective, their reliance on frequent online interactions poses risks and costs in real-world applications~\cite{kiyohara2021accelerating}. Offline RL addresses this by learning policies from static datasets, with methods like BCQ~\cite{fujimoto2019off}, CQL~\cite{kumar2020conservative}, and IQL~\cite{kostrikovoffline} offering robust solutions for stable continuous control, mitigating value overestimation, and reducing distributional shifts. However, their dependence on Markov Decision Processes (MDPs) limits access to prior observations and modeling long-range dependencies, critical in sequential tasks with strong temporal patterns. Decision Transformers (DT)~\cite{chen2021decision} overcome these limitations by reframing RL as sequential modeling, leveraging transformer architectures to capture historical patterns and long-term dependencies, achieving state-of-the-art offline RL performance. Extensions like CDT~\cite{liu2023constrained} further enable zero-shot constraint adaptation, balancing safety and performance without online fine-tuning. Building on DT, we propose an auto-bidding framework that aligns with DT's strengths by modeling bid adjustments as trajectory-based sequence generation, effectively capturing intricate temporal correlations in high-stakes, data-sensitive environments.

\subsection{Auto-Bidding at Online Advertising Platforms}

Auto-bidding plays a vital role in managing large-scale ad auctions by automatically optimizing bids per impression to meet advertisers' goals. Early methods like PID~\cite{chen2011real} and OnlineLP~\cite{yu2017online} focus on rule-based approaches, using feedback loops and stochastic programming to address budget pacing and bid optimization, though they depend on simplified assumptions. To tackle the complexity of modern advertising ecosystems, RL-based solutions such as RLB~\cite{cai2017real}, USCB~\cite{he2021unified}, MAAB~\cite{wen2022cooperative}, and SORL~\cite{mou2022sustainable} enable adaptive decision-making with capabilities for handling high-dimensional states and multi-agent coordination. However, due to the risks of real-time bidding, offline RL methods like BCQ~\cite{fujimoto2019off}, CQL~\cite{kumar2020conservative}, and IQL~\cite{kostrikovoffline} have gained prominence for leveraging historical data. These approaches, while effective under MDP frameworks, struggle with modeling long-range temporal dependencies crucial for sequential bid optimization. Generative sequential modeling methods such as DiffBid~\cite{guo2024generative} and GAS~\cite{li2024gas} address these challenges using diffusion models and transformers with MCTS for improved bid trajectory generation. In this work, we propose a score-based RTG, action exploration mechanisms, and a learnable value function framework to align optimization objectives with evaluation metrics, enhance action exploration, and learn optimal strategies for improving bidding performance with Decision Transformer.

\section{Conclusion}

In this work, we propose \name to enhance DTs for offline generative auto-bidding through value-guided explorations. To accommodate complex advertising objectives, we design a customizable score-based RTG mechanism, enabling adaptive modeling of diverse optimization objectives to align with different evaluation metrics. Moreover, we integrate an action exploration mechanism with an RTG-based evaluation method to explore actions outside the offline dataset while ensuring a stability-preserving update process. To further guide the exploration and mitigate OOD risks, we employ a learnable value function to anchor RTG updates to distributionally plausible regions while allowing controlled extrapolation for strategy improvement. Extensive experiments, online deployments and NeurIPS competition~\footnotemark[1] demonstrate the effectiveness of our \name framework in enhancing the adaptability and performance of auto-bidding strategies, providing a versatile solution for optimizing digital advertising campaigns in dynamic environments.

\begin{acks}
This research was partially supported by Kuaishou, Research Impact Fund (No.R1015-23), and Collaborative Research Fund (No.C1043-24GF).
\end{acks}

\normalem
\bibliographystyle{ACM-Reference-Format}
\bibliography{bibnew}
% \input{6appendix}
% \clearpage

\end{document}